\documentclass[sigconf]{acmart}

\usepackage{booktabs} 
\usepackage{multirow}
\usepackage{color}

\AtBeginDocument{%
  \providecommand\BibTeX{{%
    \normalfont B\kern-0.5em{\scshape i\kern-0.25em b}\kern-0.8em\TeX}}}

\copyrightyear{2020} 
\acmYear{2020} 
\setcopyright{acmcopyright}\acmConference[SIGIR '20]{Proceedings of the 43rd International ACM SIGIR Conference on Research and Development in Information Retrieval}{July 25--30, 2020}{Virtual Event, China}
\acmBooktitle{Proceedings of the 43rd International ACM SIGIR Conference on Research and Development in Information Retrieval (SIGIR '20), July 25--30, 2020, Virtual Event, China}
\acmPrice{15.00}
\acmDOI{10.1145/3397271.3401443}
\acmISBN{978-1-4503-8016-4/20/07}

\begin{document}
\fancyhead{}

\title{Entire Space Multi-Task Modeling via Post-Click Behavior Decomposition for Conversion Rate Prediction}


\author{Hong Wen}
\authornote{Both authors contributed equally to this paper.}
\affiliation{%
  \institution{Alibaba Group}
  \city{Hangzhou, Zhejiang, China 311121}
}
\email{qinggan.wh@alibaba-inc.com}

\author{Jing Zhang}
\authornotemark[1]
\affiliation{%
  \institution{The University of Sydney}
  \city{Darlington NSW 2008, Australia}
}
\email{jing.zhang1@sydney.edu.au}

\author{Yuan Wang}
\affiliation{%
  \institution{Alibaba Group}
  \city{Hangzhou, Zhejiang, China 311121}
}
\email{wy175696@alibaba-inc.com}

\author{Fuyu Lv}
\affiliation{%
  \institution{Alibaba Group}
  \city{Hangzhou, Zhejiang, China 311121}
}
\email{fuyu.lfy@alibaba-inc.com}

\author{Wentian Bao}
\affiliation{%
  \institution{Alibaba Group}
  \city{Hangzhou, Zhejiang, China 311121}
}
\email{wentian.bwt@alibaba-inc.com}

\author{Quan Lin, Keping Yang}
\affiliation{%
  \institution{Alibaba Group}
  \city{Hangzhou, Zhejiang, China 311121}
}
\email{{tieyi.lq, shaoyao}@taobao.com}



\begin{abstract}
Recommender system, as an essential part of modern e-commerce, consists of two fundamental modules, namely Click-Through Rate (CTR) and Conversion Rate (CVR) prediction. While CVR has a direct impact on the purchasing volume, its prediction is well-known challenging due to the Sample Selection Bias (SSB) and Data Sparsity (DS) issues. Although existing methods, typically built on the user sequential behavior path ``impression$\to$click$\to$purchase'', is effective for dealing with SSB issue, they still struggle to address the DS issue due to rare purchase training samples. Observing that users always take several purchase-related actions after clicking, we propose a novel idea of post-click behavior decomposition. Specifically, disjoint purchase-related Deterministic Action (DAction) and Other Action (OAction) are inserted between click and purchase in parallel, forming a novel user sequential behavior graph ``impression$\to$click$\to$D(O)Action$\to$purchase''. Defining model on this graph enables to leverage all the impression samples over the entire space and extra abundant supervised signals from D(O)Action, which will effectively address the SSB and DS issues together. To this end, we devise a novel deep recommendation model named Elaborated Entire Space Supervised Multi-task Model ($ESM^{2}$). According to the conditional probability rule defined on the graph, it employs multi-task learning to predict some decomposed sub-targets in parallel and compose them sequentially to formulate the final CVR. Extensive experiments on both offline and online environments demonstrate the superiority of $ESM^{2}$ over state-of-the-art models. The source code and dataset will be released.

\end{abstract}

\ccsdesc[500]{Computer systems organization~Neural networks}
\ccsdesc[500]{Information systems~Recommender systems}

\keywords{Recommender System, Entire Space Multi-Task Learning, Post-Click Behavior Decomposition, Conversion Rate Prediction}


\maketitle

\section{Introduction}
\begin{figure}[h]
  \centering
  \includegraphics[width=\linewidth]{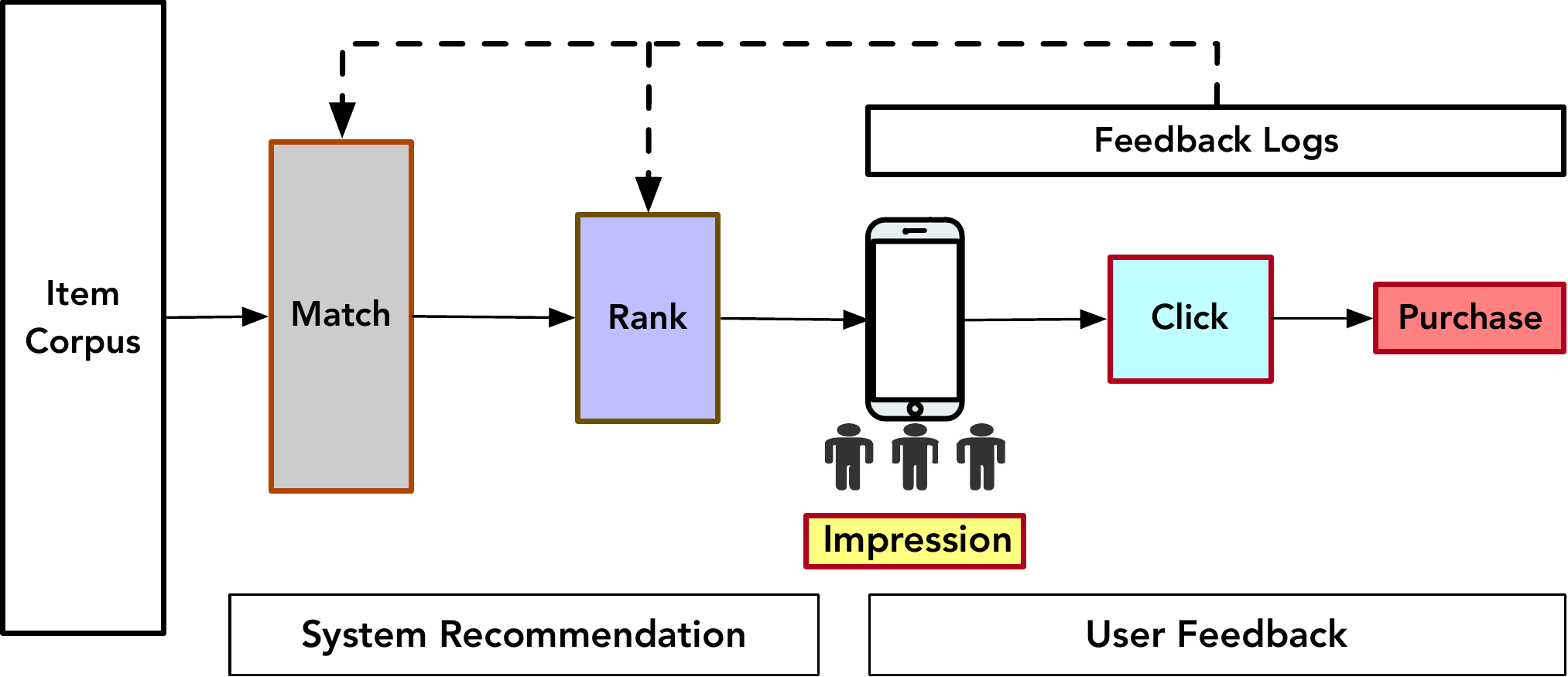}
  \caption{A diagram of online recommendation in e-commerce platform, comprising of two fundamental components, $i.e.$, system recommendation and user feedback.}
  
  \label{fig:rs_architecture}
\end{figure}

Discovering valuable products or services from massive available options on the Internet for users has become a fundamental functionality in modern online applications such as e-commerce \cite{ni2018perceive, chen2019behavior, sun2019bert4rec, wen2019multi}, social networking \cite{golbeck2006filmtrust, naruchitparames2011friend}, advertising \cite{zhou2019deep, zhou2018deep}, $etc$. Recommender System (RS) can serve this role to provides accurate, timely, and personalized services to users \cite{zhu2018learning, feng2019deep, lv2019sdm}. Figure~\ref{fig:rs_architecture} shows the architecture of online recommendation in e-commerce platform. It consists of two phases, $i.e.$, system recommendation and user feedback. After analyzing users' long and short-term behaviors, RS first recalls a large number of related items. Then, they are ranked and exposed to users according to several ranking metrics, $e.g.$, Click-Through Rate (CTR) \cite{zhou2019deep, zhou2018deep}, Conversion Rate (CVR) \cite{wen2019multi, ma2018entire} , $etc.$ Next, when going through the recommended items, users may click on and eventually purchase the interested ones, indicating a typical user sequential behavior path ``impression$\to$click$\to$purchase'' for e-commerce transaction \cite{ma2018entire}. These feedback from users are collected by RS and used to estimate more accurate ranking metrics, which are indeed very crucial for generating high-quality recommendations in turn. In this paper, we focus on the post-click CVR estimation task.

\begin{figure}[ht]
  \centering
  \includegraphics[width=0.8\linewidth]{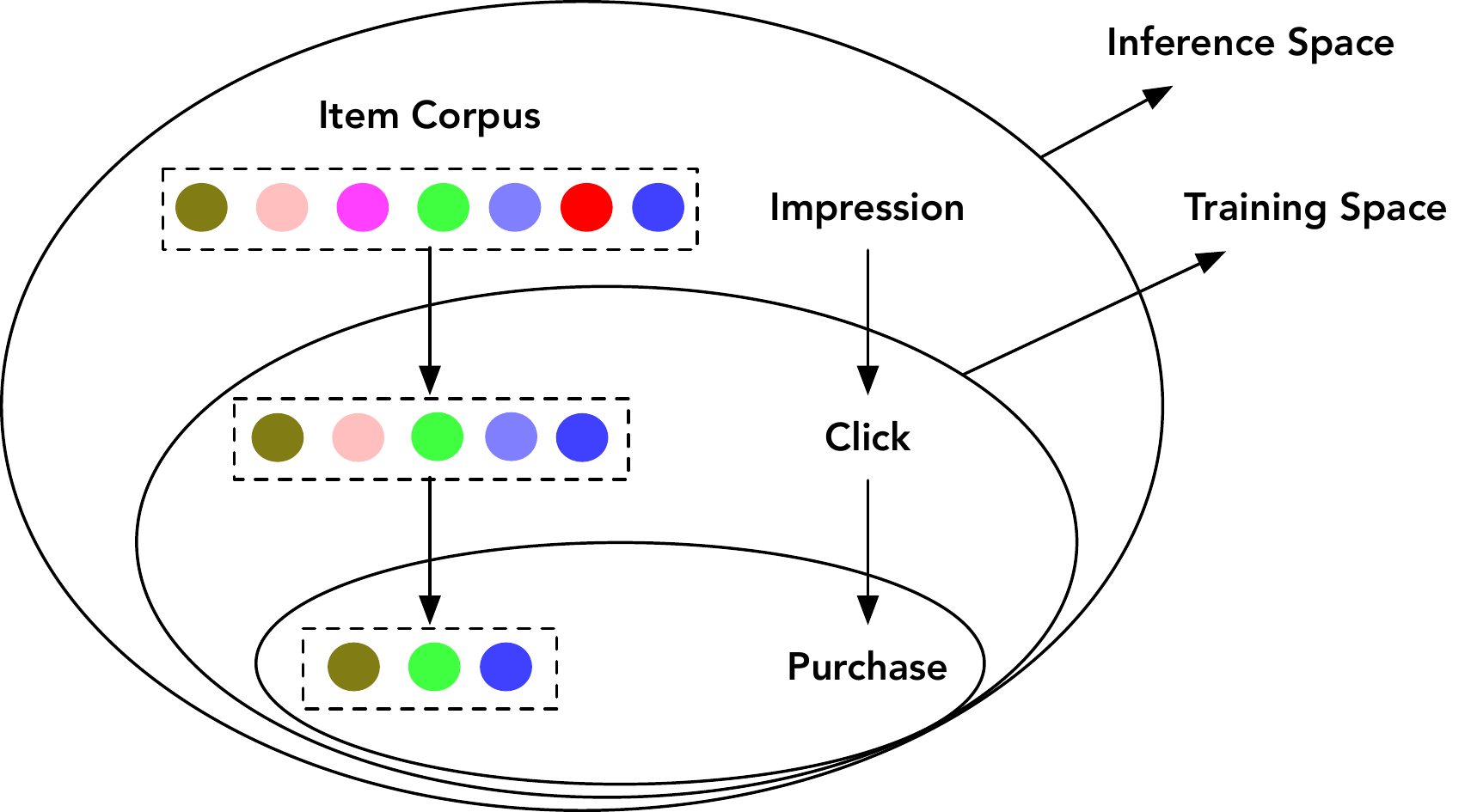}
  \caption{Illustration of sample selection bias problem in conventional CVR prediction, where training space only composes of clicked samples, while inference space is the entire space for all impression samples. And data volume gradually decreased from impression to purchase.}
  \label{fig:ssb}
\end{figure}

However, two critical issues in the CVR estimation makes the task quite challenging, $i.e.$, Sample Selection Bias (SSB) \cite{zadrozny2004learning} and Data Sparsity (DS) \cite{lee2012estimating}. 
SSB refers to the systematic difference of data distributions between training space and inference space, $i.e.$, conventional CVR models are trained only on clicked samples while being used for inference on all impression samples. Intuitively, clicked samples are only a very small portion of the impression samples and are biased by user self-selection (such as clicking). Therefore, when the CVR model serving online, the SSB issue will degrade its performance. Besides, due to the relatively rare clicking samples compared with impressions, the number of training samples from the sequential behavior path ``click$\to$purchase'' is insufficient to fit the large parameter space of CVR task, which results in the DS problem. As illustrated in Figure~\ref{fig:ssb}, how to deal with the SSB and DS problems is crucial for developing an efficient industrial-level recommender system.

Several studies have been carried out to tackle these challenges \cite{pan2008one, lee2012estimating, zhang2016bid, weiss2004mining,ma2018entire}. For example, Ma $et~al.$ propose a new model named Entire Space Multi-Task Model (ESMM) \cite{ma2018entire}, which defines CVR task on the user sequential behavior path ``impression$\to$click$\to$purchase'' via multi-task learning framework. It is trained with all impression samples over the entire space for two auxiliary tasks namely post-view CTR and post-view click-through conversion rate (CTCVR). Therefore, the derived CVR from CTR and CTCVR is also applicable in the same entire space when inferring online, thus addressing the SSB issue effectively. Besides, an auxiliary CTR network with rich labeled samples shares the same feature representation with the CVR network, helping to alleviate the DS issue. Although ESMM achieves better performance than conventional methods by dealing with the SSB and DS issues simultaneously, it still struggles to alleviate the DS issue due to the rare purchase training samples, $i.e.$, less than 0.1\% of impression behaviors converts to purchase according to the large scale real transaction logs from our e-commerce platform.

After a detailed analysis of the logs, we observe that users always take some purchase-related actions after clicking. For example, users may add the preferred items to their shopping cart (or wish list) instead of immediately purchases due to some reasons ($i.e.$, waiting for a discount). Besides, these actions are indeed more abundant than purchase actions. Motivated by this, we propose a novel idea of post-click behavior decomposition. Specifically, disjoint purchase-related Deterministic Action (DAction) and Other Action (OAction) are inserted between click and purchase in parallel, forming a novel user sequential behavior graph ``impression$\to$click$\to$D(O)Action$\to$purchase'', where the task relationship is explicitly defined by the conditional probability. Besides, defining model on this graph enables to leverage all impression samples over the entire space and extra abundant supervisory signals from post-click behaviors, efficiently addressing the SSB and DS issues.

In this paper, we resort to deep neural networks to embody the above idea. Specifically, we propose a novel deep neural recommendation model named Elaborated Entire Space Supervised Multi-task Model ($ESM^{2}$), which consists of three modules: 1) a shared embedding module (SEM), 2) a decomposed prediction module (DPM), and 3) a sequential composition module (SCM). First, SEM embeds a one-hot feature vector of ID features into dense representation through a linear fully connected layer. Then, these embeddings are fed into the subsequent DPM, where individual prediction network estimates the probabilities of decomposed sub-targets in parallel by employing multi-task learning on all the impression samples over the entire space. Finally, SCM composes the final CVR as well as some auxiliary probabilities sequentially according to the conditional probability rule defined on the graph. Multiple losses defined on some sub-paths of the graph are used to supervise the training of $ESM^{2}$.

The main contributions of this paper are summarized as follows:

$\bullet$ To the extent of our knowledge, we are the first to introduce the idea of post-click behavior decomposition to model CVR over the entire space. The explicit decomposition results in a novel user sequential behavior graph ``impression$\to$click$\to$D(O)Action$\to$purchase''.

$\bullet$ We propose a novel deep neural recommendation method named $ESM^{2}$, which models CVR prediction and auxiliary tasks simultaneously in a multi-task learning framework according to the conditional probability rule defined on the user behavior graph. $ESM^{2}$ can address the SSB and DS issues efficiently by harvesting the abundant post-click action data with labels.

$\bullet$ Our model achieves better performance on the real-world offline dataset than representative state-of-the-art methods. We also deploy it in our online recommender system and achieve significant improvement, confirming its value in industrial applications. 

The rest of this paper is organized as follows. Section 2 presents a brief survey of related work, followed by the details of the proposed model in Section 3. Experiment results and analysis are presented in Section 4. Finally, we conclude the paper in Section 5.

\section{Related work}
Our proposed method specifically tackles the conversion rate prediction problem by employing the multi-task learning framework over the entire space. Therefore, we briefly review the most related work from the following two aspects: 1) conversion rate prediction and 2) multi-task learning.

\begin{figure*}
  \centering
  \includegraphics[width=\linewidth]{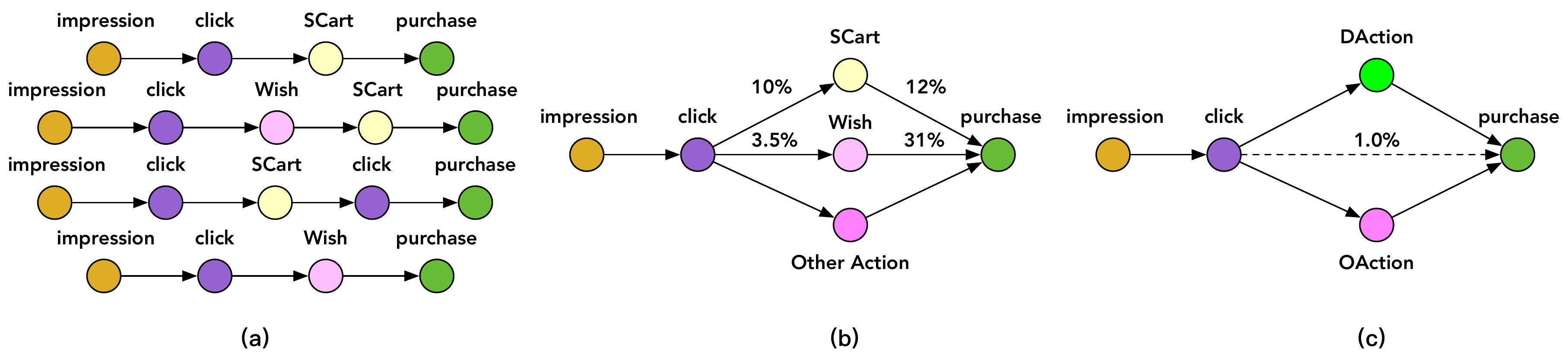}
  \caption{Illustration of the proposed user sequential behavior graph based on post-click behavior decomposition. (a) The multiple paths from impression to purchase after distinguishing post-click behaviors, such as ``impression$\to$click$\to$SCart$\to$purchase''. (b) A digraph is used to describe the simplified purchasing process, where the numbers above edges represent the sparsity of different paths. (c) Several specific purchase-related post-click actions are merged into a single node, $i.e.$, DAction, which also inherits their supervisory signals. OAction represents other cases except DAction.}
  \label{fig:multi_path}
\end{figure*}

\textbf{Conversion Rate Prediction: } Conversion rate prediction is a key component of many online applications, such as search engines \cite{dupret2008user, zhang2014sequential}, recommender systems \cite{guo2017deepfm,qu2016product} and online advertising \cite{graepel2010web, he2014practical}. However, there are few literatures directly proposed for CVR tasks \cite{lu2017practical, wen2019multi, yang2016large}, regardless of recent prosperous development of CTR methods \cite{effendi2017click,zhou2018deep,zhou2019deep, xiao2017attentional}. Indeed, CVR modeling is very challenging since conversions are extremely rare events that only a very small portion of impression items are eventually being clicked and bought. Recently, the deep neural network has achieved significant progress in many areas including recommender systems due to its remarkable ability in feature representation and end-to-end modeling \cite{hinton2006fast, graves2013speech, krizhevsky2012imagenet, shen2014latent, feng2019deep, lv2019sdm}. In this paper, we also adopt a deep neural network to model the conversion rate prediction task. In contrast to the above methods, we derive a new user sequential behavior graph ``impression$\to$click$\to$D(O)Action$\to$purchase'' based on a novel idea of post-click behavior decomposition. According to the conditional probability rule defined on the graph, our network structure is specifically devised to predict several decomposed sub-targets in parallel and compose them sequentially to formulate the final CVR.

\textbf{Multi-Task Learning: } 
Due to the temporal multi-stage nature of users' purchasing behavior, $e.g.$, impression, click, and purchase, prior work attempts to formulate the conversion rate prediction task by a multi-task learning framework. For example, Hadash $et~al.$ propose a multi-task learning-based recommender system by modeling the ranking and rating prediction tasks simultaneously \cite{hadash2018rank}. Ma $et~al.$ propose a multi-task learning approach named multi-gate mixture-of-experts to explicitly learn the task relationship from data \cite{ma2018modeling}. Gao $et~al.$ propose a neural multi-task recommendation model to learn the cascading relationship among different types of behaviors \cite{gao2019neural}. In contrast, we model the CTR and CVR tasks simultaneously by associating with users' sequential behavior graph, where the task relationship is explicitly defined by the conditional probability (See Section \ref{sec:method}). Ni $et~al.$ propose to learn universal user representations across multiple tasks for more effective personalization \cite{ni2018perceive}. We also explore such an idea by sharing embedded features across different tasks. Recently, Ma $et~al.$ propose an entire space multi-task model (ESMM) for CVR prediction \cite{ma2018entire}. It adds the CTR task and CTCVR task as an auxiliary to the main CVR task. Our method is partially inspired by ESMM but has the following significant difference: we propose a novel idea of post-click behavior decomposition to reformulate a novel user sequential behavior graph ``impression$\to$click$\to$D(O)Action$\to$purchase''. Defining model on this graph enables to formulate the final CVR as well as some auxiliary tasks together. It can leverage all the impression samples over the entire space and the abundant supervisory signals from users' post-click behaviors, which are highly relevant to the purchase behaviors, consequently addressing the SSB and DS issue simultaneously.

\section{Proposed method}
\label{sec:method}
\subsection{Motivation}
\label{subsec:motivation}
In practice, from an item being displayed to it being purchased successfully, we identify that there may exist multiple kinds of sequential actions a user could choose to do. For example, after clicking one interested item, a user may directly purchase it without any hesitation, or add it to the shopping cart and then make the purchase eventually. These behavior paths are shown in Figure~\ref{fig:multi_path}(a). We can simplify and group these paths according to several predefined specific purchase-related post-click actions, $i.e.$, adding to Shopping Cart (SCart) and adding to Wish list (Wish), as shown in Figure~\ref{fig:multi_path}(b). Based on our data analysis of online real-world logs, we found that only 1\% of clicked behaviors are converted to purchase eventually, indicating rare purchase training samples. However, the data volume of several post-click actions like SCart and Wish are much larger than purchase. For example, 10\% will be added to the shopping cart given clicked behaviors. Besides, these post-click actions are highly relevant to the final purchase action, $e.g.$, 12\% (or 31\%) will be bought eventually after they have been added to the shopping cart (or wish list). How can we leverage the larger volume of post-click behaviors to benefit CVR prediction in some manner, regarding their high relevance to purchase?

Intuitively, one solution is to model these purchase-related post-click actions along with purchase into a multi-task prediction framework. The key is how to formulate them properly since they have explicit sequential correlations, $e.g.$, the purchase action probably conditioned on the SCart or Wish action. To this end, we define a single node named Deterministic Action (DAction) to merge these predefined specific purchase-related post-click actions, such as SCart and Wish, as shown in Figure~\ref{fig:multi_path}(c). DAction has two properties: 1) it is highly relevant to the purchase action and 2) it has abundant deterministic supervisory signals from users' feedback, $e.g.$, 1 for taking some specific actions ($i.e.$, adding to shopping cart or wish list after clicking) and 0 for none. We also add a node named Other Action(OAction) between click and purchase to deal with other post-click behaviors except DAction. In this way, the conventional behavior path ``impression$\to$click$\to$purchase'' becomes to a novel elaborated user sequential behavior graph ``impression$\to$click$\to$D(O)Action$\to$purchase'', as shown in Figure~\ref{fig:multi_path}(c). Defining model on this graph enables to leverage all the impression samples over the entire space and extra abundant supervisory signals from D(O)Action, which will circumvent the SSB and DS issues efficiently. We call this novel idea as post-click behavior decomposition.

\subsection{Conditional probability decomposition}
\label{subsec:probabilityDecompositionCVR}
In this section, we present the conditional probability decomposition of CVR as well as related auxiliary tasks according to the digraph defined in Figure~\ref{fig:multi_path}(c). 
First, the probability of post view click-through rate of an item $x_i$, denoted as $p_i^{ctr}$, is defined as the conditional probability of being clicked given that it has been viewed, which depicts the path ``impression$\to$click'' in the digraph. Mathematically, it can be written as:
\begin{equation}
    p_i^{ctr} = p\left( {{c_i} = 1\left| {{v_i} = 1} \right.} \right) \buildrel \Delta \over = y_{1i},
\label{eq:pctr}
\end{equation}
where ${c_i} \in C$ denotes whether the $i^{th}$ item $x_i$ is being clicked, ${c_i} \in \left\{ {0,1} \right\}$, $C$ is the label spaces of all the items being clicked or not, $i \in \left[ {1,N} \right]$ and $N$ is the number of items. Similarly, ${v_i} \in V$ denotes whether the $i^{th}$ item $x_i$ is being viewed ($i.e.$, impression), ${v_i} \in \left\{ {0,1} \right\}$, $V$ is the label spaces of all the items being viewed or not. $y_{1i}$ is a surrogate symbol for simplicity.

Then, the probability of click-through DAction conversion rate of an item $x_i$, denoted as $p_i^{ctavr}$, is defined as the conditional probability of being taken DAction given that it has been viewed, which depicts the path ``impression$\to$click$\to$DAction'' in the digraph. Mathematically, it can be written as:
\begin{equation}
\begin{aligned}
 p_i^{ctavr} &= p\left( {{a_i} = 1\left| {{v_i} = 1} \right.} \right) \\
  &= \sum\limits_{{c_i} \in \left\{ {0,1} \right\}} {p\left( {{a_i} = 1\left| {{v_i} = 1} \right.,{c_i}} \right)p\left( {{c_i}\left| {{v_i} = 1} \right.} \right)}  \\
  &= p\left( {{a_i} = 1\left| {{v_i} = 1} \right.,{c_i} = 0} \right)p\left( {{c_i} = 0\left| {{v_i} = 1} \right.} \right) \\
  & \quad + p\left( {{a_i} = 1\left| {{v_i} = 1} \right.,{c_i} = 1} \right)p\left( {{c_i} = 1\left| {{v_i} = 1} \right.} \right) \\
  &= {y_{2i}}{y_{1i}} \\
 \end{aligned},
\label{eq:pctavr}
\end{equation}
where ${a_i} \in A$ denotes whether the $i^{th}$ item $x_i$ is being taken some specific actions defined in Section~\ref{subsec:motivation}, ${a_i} \in \left\{ {0,1} \right\}$, $A$ is the label spaces of all the items being taken some specific actions or not. ${y_{2i}} = p\left( {{a_i} = 1\left| {{v_i} = 1} \right.,{c_i} = 1} \right)$, depicting the path ``click$\to$DAction'', is a surrogate symbol for simplicity as $y_{1i}$. It is trivial that ${y_{2i}} = p\left( {{a_i} = 1\left| {{c_i} = 1} \right.} \right)$ since all the samples are impression samples ($i.e.$, $v_i = 1$). It is noteworthy that Eq.~\eqref{eq:pctavr} holds due to the fact that no action occurs without being clicked, $i.e.$, $p\left( {{a_i} = 1\left| {{v_i} = 1} \right.,{c_i} = 0} \right){\rm{ = }}0$.

Next, the probability of conversion rate of an item $x_i$, denoted as $p_i^{cvr}$, is defined as the conditional probability of being bought given that it has been clicked, which depicts the paths ``click$\to$D(O)Action$\to$purchase'' in the digraph. Mathematically, it can be written as:
\begin{equation}
\begin{aligned}
 p_i^{cvr} &= p\left( {{b_i} = 1\left| {{c_i} = 1} \right.} \right) \\
  &= \sum\limits_{{a_i} \in \left\{ {0,1} \right\}} {p\left( {{b_i} = 1\left| {{c_i} = 1} \right.,{a_i}} \right)p\left( {{a_i}\left| {{c_i} = 1} \right.} \right)} \\
  &= p\left( {{b_i} = 1\left| {{c_i} = 1} \right.,{a_i} = 0} \right)p\left( {{a_i} = 0\left| {{c_i} = 1} \right.} \right) \\
  & \quad + p\left( {{b_i} = 1\left| {{c_i} = 1} \right.,{a_i} = 1} \right)p\left( {{a_i} = 1\left| {{c_i} = 1} \right.} \right) \\
  & \buildrel \Delta \over = {y_{4i}}\left( {1 - {y_{2i}}} \right) + {y_{2i}}{y_{3i}} \\
 \end{aligned},
\label{eq:pcvr}
\end{equation}
where ${b_i} \in B$ denotes whether the $i^{th}$ item $x_i$ is being bought, ${b_i} \in \left\{ {0,1} \right\}$, $B$ is the label spaces of all the items being bought or not. ${y_{3i}} = p\left( {{b_i} = 1\left| {{c_i} = 1} \right.,{a_i} = 1} \right)$, ${y_{4i}} = p\left( {{b_i} = 1\left| {{c_i} = 1} \right.,{a_i} = 0} \right)$ are some surrogate symbols for simplicity as $y_{1i}$. ${y_{3i}}$ or ${y_{4i}}$ depicts the path ``DAction$\to$purchase'' or ``OAction$\to$purchase'' in the digraph, respectively.

The probability of click-through conversion rate of an item $x_i$, denoted as $p_i^{ctcvr}$, is defined as the conditional probability of being bought given that it has been viewed, which depicts the complete graph ``impression$\to$click$\to$D(O)Action$\to$purchase'' in the digraph. Mathematically, it can be written as:
\begin{equation}
\begin{aligned}
 p_i^{ctcvr} &= p\left( {{b_i} = 1\left| {{v_i} = 1} \right.} \right) \\
  &= \sum\limits_{{c_i}} {p\left( {{b_i} = 1\left| {{v_i} = 1} \right.,{c_i}} \right)p\left( {{c_i}\left| {{v_i} = 1} \right.} \right)}  \\
  &= \sum\limits_{{c_i}} {\sum\limits_{{a_i}} {p\left( {{b_i} = 1\left| {{v_i}} \right.,{c_i},{a_i}} \right)p\left( {{a_i}\left| {{v_i}} \right.,{c_i}} \right)p\left( {{c_i}\left| {{v_i}} \right.} \right)} }  \\
  &= {y_{4i}}\left( {1 - {y_{2i}}} \right){y_{1i}} + {y_{3i}}{y_{2i}}{y_{1i}} \\
  &= {y_{1i}}\left( {{y_{4i}}\left( {1 - {y_{2i}}} \right) + {y_{3i}}{y_{2i}}} \right) \\
 \end{aligned}.
\label{eq:pctcvr}
\end{equation}
Here, we use $v_i$ to replace ${{v_i} = 1}$ in the third equality for simplicity without causing any ambiguity. It is noteworthy that the fourth equality holds due to the fact that no items will be bought without being clicked, $i.e.$, $p\left( {{b_i} = 1\left| {{v_i} = 1} \right.,{c_i} = 0,{a_i}} \right) $ equals to zero, $\forall {a_i} \in \left\{ {0,1} \right\}$. Indeed, Eq.~\eqref{eq:pctcvr} can be derived by decomposing the graph ``impression$\to$click$\to$D(O)Action$\to$purchase'' into ``impression$\to$click'' and ``click$\to$D(O)Action$\to$purchase'', and integrating Eq.~\eqref{eq:pctr} and Eq.~\eqref{eq:pcvr} together according to the chain rule, $i.e.,$ $p_i^{ctcvr}=p_i^{ctr}*p_i^{cvr}$.

\begin{figure*}
  \centering
  \includegraphics[width=\linewidth]{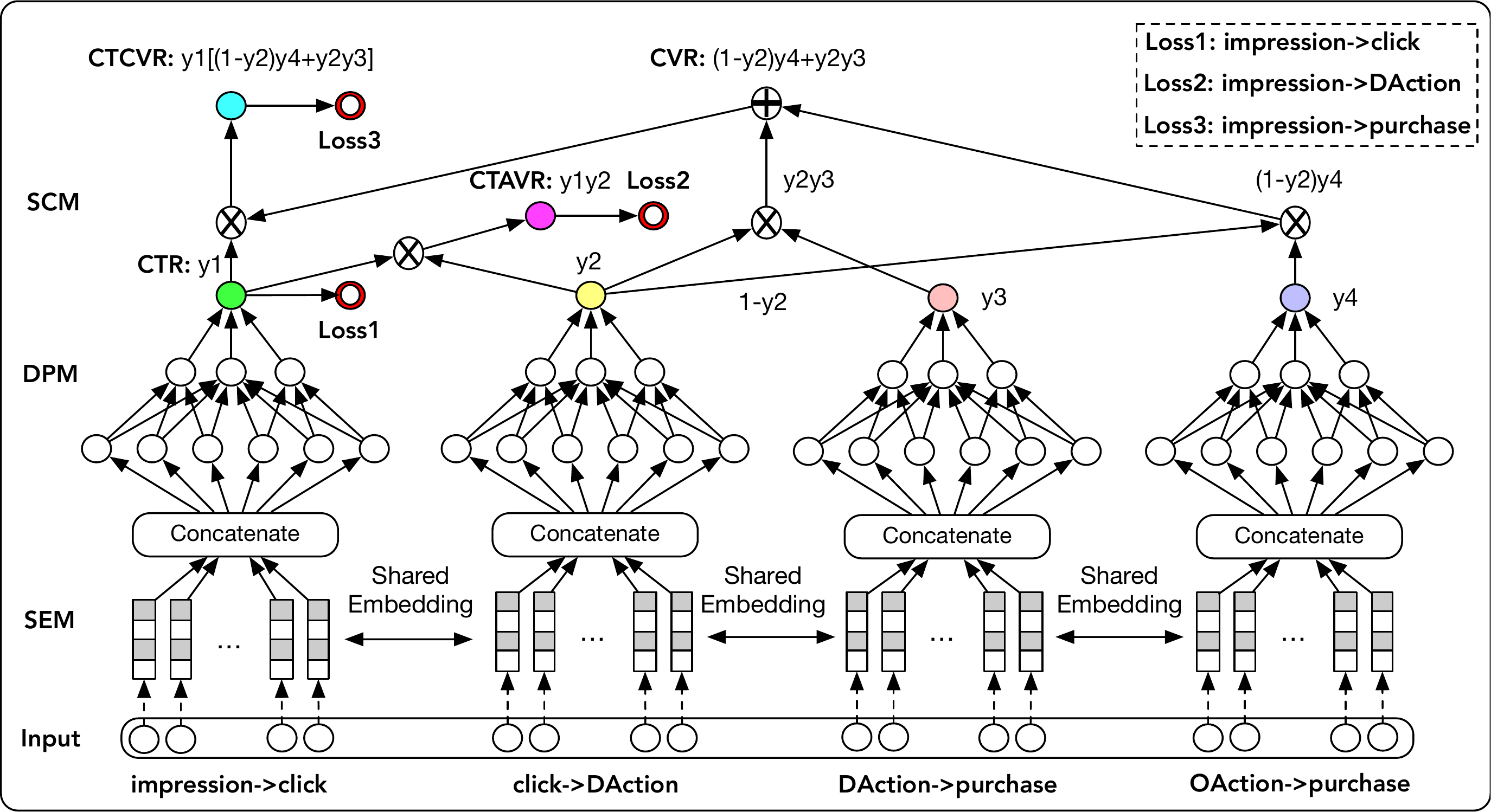}
  \caption{ 
  The diagram of $ESM^2$ model over the entire space, which consists of three key modules: SEM, DPM and SCM. SEM embeds sparse features into dense representation. DPM predicts the probabilities of decomposed targets. SCM integrates them together sequentially to calculate the final CVR as well as other related auxiliary tasks namely CTR, CTAVR, and CTCVR.
  }
  \label{fig:model}
\end{figure*}

\subsection{Elaborated entire space supervised multi-task model}
\label{subsec:ESM}
From Eq.~\eqref{eq:pctr}$ \sim$ Eq.~\eqref{eq:pctcvr}, we can see that $p_i^{ctr}$, $p_i^{ctavr}$, and $p_i^{ctcvr}$ can be derived from four hidden probability variables $y_{1i}$, $y_{2i}$, $y_{3i}$, and $y_{4i}$, which represents the conditional probabilities over some sub-paths in the graph, $i.e.$, ``impression$\to$click'', ``click$\to$DAction'', ``DAction$\to$purchase'' and ``OAction$\to$purchase''. 
On the one hand, these four sub-targets are defined over the entire space and can be predicted using all the impression samples. Taking $y_{2i}$ as an example, training $y_{2i}$ directly with only clicked samples suffers from the SSB issue. Indeed, $y_{2i}$ is an intermediate variable derived from $p_i^{ctr}$ and $p_i^{ctavr}$ according to Eq.~\eqref{eq:pctavr}. Since both $p_i^{ctr}$ and $p_i^{ctavr}$ are modeled over the entire space with all impression samples, the derived $y_{2i}$ is also applicable over the entire space, therefore, no SSB in our model. On the other hand, ground truth labels of $p_i^{ctr}$, $p_i^{ctavr}$, and $p_i^{ctcvr}$ are available given users' logs, which can be used to supervise these sub-targets. Therefore, an intuitive way is to model them simultaneously by employing a multi-task learning framework. To this end, we propose a novel deep neural recommendation model named Elaborated Entire Space Supervised Multi-task Model ($ESM^2$) for CVR prediction. $ESM^2$ gets its name since 1) $p_i^{ctr}$, $p_i^{ctavr}$, and $p_i^{ctcvr}$ are modeled over the entire space and predicted using all the impression samples; 2) the derived $p_i^{cvr}$ from Eq.~\eqref{eq:pcvr} also benefits from the entire space multi-task modeling, which will be validated in the experiment part. $ESM^2$ consists of three key modules: 1) a shared embedding module, 2) a decomposed prediction module, and 3) a sequential composition module. We present each of them in detail as follows.

\textbf{Shared Embedding Module (SEM):} First, we devise a shared embedding module to embed all the sparse ID features and dense numerical features coming from user field, item field, and user-item cross field. The user features include users' ID, ages, genders and purchasing powers, $etc.$ The item features include items' ID, prices, accumulated CTR and CVR from historical logs, $etc.$ The user-item features include users' historical preference scores on items, $etc.$ Dense numerical features are first discretized based on their boundary values and then represented as one-hot vectors. Here, we use ${f_i} = \left\{ {{f_{ij,}}\forall j \in {\Lambda _f}} \right\}$ to denote the one-hot features of the $i^{th}$ training sample, where ${{\Lambda _f}}$ denotes the index set of all kinds of features. Due to the sparseness nature of one-hot encoding, we employ linear fully connected layers to embed them into dense representation, which can be formulated as:
\begin{equation}
    g_{ij}=P_{\theta_j}^{T}f_{ij},
   \label{eq:sem}
\end{equation}
where $P_{\theta_j}$ denotes the embedding matrix for the $j^{th}$ kind of features, $\theta_j$ represents the network parameters.

\textbf{Decomposed Prediction Module (DPM):} Then, once all the feature embeddings are obtained, they are concatenated together, fed into several decomposed prediction modules, and shared by each of them. Each prediction network in DPM estimates the probability of the decomposed target on the path ``impression$\to$click'', ``click$\to$DAction'', ``DAction$\to$purchase'', ``OAction$\to$purchase'', respectively. In this paper, we employ Multi-Layer Perception (\emph{MLP}) as the prediction network. All the non-linear activation function is \emph{ReLU} except the output layer, where we use a \emph{Sigmoid} function to map the output into a probability taking real value from 0 to 1. Mathematically, it can be formulated as:
\begin{equation}
    y_{ki}=\sigma \left ( \varphi _{\vartheta_{k} }^{k}\left ( g_{i} \right ) \right ),
   \label{eq:dpm}
\end{equation}
where $\sigma$ denotes the \emph{Sigmoid} function, $\varphi _{\vartheta_{k} }^{k}$ denotes the mapping function learned by the $k$th MLP, $\vartheta_{k}$ denotes its network parameters. For example, as shown in the first \emph{MLP} in Figure~\ref{fig:model}, it output the estimated probability $y_1$, which is indeed the post-view click-through rate.

\textbf{Sequential Composition Module (SCM):} Finally, we devise a sequential composition module to compose the above predicted probabilities sequentially according to Eq.~\eqref{eq:pctr} $\sim$ Eq.\eqref{eq:pctcvr} to calculate the conversion rate $p^{cvr}$ and some auxiliary targets including the post-view click-through rate $p^{ctr}$, click-through DAction conversation rate $p^{ctavr}$, and click-through conversion rate $p^{ctcvr}$, respectively. As shown in the top part of Figure~\ref{fig:model}, SCM is a parameter-free feed forward neural network which represents the underlying conditional probabilities defined by the purchasing decision digraph in Figure~\ref{fig:multi_path}.

\textbf{Remarks:} 1) All the tasks share the same embedding, making them be trained with all impression samples, $i.e.$, they are modeled over the entire space, resulting in no SSB issue during the inference phase; 2) the lightweight DPM is strictly regularized by the shared embedding module, which makes up the majority of the trainable parameters; and 3) our model suggests an efficient network design, where SEM can run in parallel, leading to low latency when deployed online.

\subsection{Training objective}
\label{subsec:objective}
We use $S=\left \{(c_{i},a_{i},b_{i};f_{i}) \right \}|_{i=1}^{N}$ to denote the training set, where $c_{i}$, $a_{i}$, $b_{i}$, represent the ground truth label whether the $i^{th}$ impression sample is being clicked, taken deterministic actions, and bought. Then, we can define the joint post-view click-through probability of all training samples as follows:
\begin{equation}
    p^{ctr}=\prod_{i\in C_{+}}{p}_{i}^{ctr}\prod_{j\in C_{-}}\left (1-{p}_{j}^{ctr} \right ),
\label{eq:jointPctr}
\end{equation}
where $C_{+}$ and $C_{-}$ denote the positive and negative samples in the label space $C$, respectively. After taking negative logarithm on Eq.\eqref{eq:jointPctr}, we obtain the \emph{logloss} of $p^{ctr}$, which is widely used in recommender systems, $i.e.$,
\begin{equation}
    L_{ctr}=-\sum_{i\in C_{+}}log{p}_{i}^{ctr}-\sum_{j\in C_{-}}log\left (1-{p}_{j}^{ctr}\right ).
    \label{eq:lossCtr}
\end{equation}
Similarly, we can obtain the loss function of $p^{ctavr}$ and $p^{ctcvr}$as follows:
\begin{equation}
    L_{ctavr}=-\sum_{i\in A_{+}}log{p}_{i}^{ctavr}-\sum_{j\in A_{-}}log\left (1-{p}_{j}^{ctavr}\right ),
    \label{eq:lossCtavr}
\end{equation}
and
\begin{equation}
    L_{ctcvr}=-\sum_{i\in B_{+}}log{p}_{i}^{ctcvr}-\sum_{j\in B_{-}}log\left (1-{p}_{j}^{ctcvr}\right ).
    \label{eq:lossCtcvr}
\end{equation}

The final training objective to be minimized is defined as:
\begin{equation}
L\left( \Theta \right ) = w_{ctr} \times L_{ctr} + w_{ctavr} \times L_{ctavr} + w_{ctcvr} \times L_{ctcvr},
 \label{eq:loss}
\end{equation}
where $\Theta =\left \{ \theta_j, \forall j \in \Lambda_f \right \}\cup \left \{ \vartheta_i,i=1,2,3,4  \right \}$ denotes all the network parameters in $ESM^2$. $w_{ctr}$, $w_{ctavr}$, $w_{ctcvr}$ are loss weights of $L_{ctr}$, $L_{ctavr}$,  $L_{ctcvr}$, which are set to 1 in this paper, respectively.

\textbf{Remarks:} 1) Adding intermediate losses to supervise the decomposed sub-tasks can efficiently leverage the abundant labeled data from post-click behaviors, making the model less affected by the DS issue; and 2) all the losses are computed from the view of entire space modeling, effectively addressing the SSB issue. 

\section{Experiments}
\label{sec:experiment}
To evaluate the effectiveness of the proposed $ESM^2$ model, we conducted extensive experiments on both the offline dataset collected from real-world e-commerce scenarios and online deployment. $ESM^2$ is compared with some representative state-of-the-art (SOTA) methods including GBDT \cite{friedman2001greedy}, DNN \cite{hinton2006fast}, DNN using over-sampling idea \cite{pan2008one} and ESMM \cite{ma2018modeling}. First, we present the evaluation settings including the dataset preparation, evaluation metrics, a brief description of these SOTA methods, and the implementation details. Then, we present the comparison results and analysis. Ablation studies are presented next, followed by the performance analysis on different post-click behaviors.

\subsection{Evaluation settings}
\label{subsec:settings}
\subsubsection{Dataset preparation}
\label{subsubsec:dataset}
We make the offline dataset by collecting the users' sequential behaviors and feedback logs \footnote{To the extent of our knowledge, there are no public datasets suited for this entire space modeling task, we will release our dataset for reproducibility and future research.} from our online e-commerce platform, which is one of the largest third-party retail platforms in the world. More than 300 million instances with user/item/user-item features and sequential feedback labels ($e.g.$, whether click, or DAction, or purchase) are filtered out. The statistics of this offline dataset are listed in Table~\ref{tab:stats}. They are further divided into the disjoint training set, validation set, and test set.

\begin{table}[htbp]
  \caption{Statistics of the offline dataset.}
  \label{tab:data}
  \begin{tabular}{llll}
    \toprule
   Category & \#User & \#Item & \#Impression  \\
    \midrule
   Number & 13,383,415 & 10,399,095 & 326,325,042 \\
    \hline
    \hline
   Category& \#Click & \#Purchase & \#DAction \\
    \hline
   Number & 20,637,192 & 226,918 & 2,501,776\\
  \bottomrule
\end{tabular}
\label{tab:stats}
\end{table}

\subsubsection{Evaluation metrics}
\label{subsubsec:metrics}
To comprehensively evaluate the effectiveness of the proposed model and compare it with SOTA methods, we adopt three widely used metrics in recommendation and advertising system, $i.e.$, Area Under Curve (AUC), GAUC \cite{zhou2018deep, zhu2017optimized} and $F_{1}$ score, where AUC reflecting the ranking ability.
\begin{equation}
    AUC = \frac{1}{{\left| {{S_ + }} \right|\left| {{S_ - }} \right|}}\sum\limits_{{x^ + } \in {S_ + }} {\sum\limits_{{x^ - } \in {S_ - }} {I\left( {\phi \left( {{x^ + }} \right) > \phi \left( {{x^ - }} \right)} \right)} },
\end{equation}
where $S_{+}$ and $S_{-}$ denote the set of positive/negative samples, respectively, $\left| {{S_ + }} \right|$ and $\left| {{S_ - }} \right|$ denote the number of samples in $S_{+}$ and $S_{-}$, $\phi \left(  \cdot  \right)$ is the prediction function, $I\left(  \cdot  \right)$ is the indicator function.

GAUC \cite{zhu2017optimized} is calculated as follows. First, all the test data are partitioned into different groups according to the individual user ID. Then, AUC is calculated in every single group. Finally, we average the weighted AUC. Mathematically, GAUC is defined as:
\begin{equation}
    GAUC=\frac{\sum_{u}w_{u} \times AUC_{u}}{\sum_{u}w_{u}},
\end{equation}
where $w_{u}$ denotes the weight for user $u$ (set as 1 for our offline evaluations). $AUC_{u}$ denotes the AUC for user $u$.

Moreover, $F_{1}$ score is defined as:
\begin{equation}
    F_{1}=\frac{2 \times P \times R}{P + R},
    \label{eq:f1}
\end{equation}
where $P$ and $R$ denote the precision and recall, $i.e.$,:
\begin{equation}
   P=\frac{TP}{TP + FP},
   \label{eq:precision}
\end{equation}
\begin{equation}
   R=\frac{TP}{TP + FN},
   \label{eq:recall}
\end{equation}
where $TP$, $FP$, and $FN$ denote the number of true positive, false positive, and false negative predictions, respectively.

\subsubsection{Brief description of comparison methods}
\label{subsubsec:sotaMethods}
The representative state-of-the-art methods are described as follows.

$\bullet$ \textbf{GBDT} \cite{friedman2001greedy}: The gradient boosting decision tree (GBDT) model follows the idea of gradient boosting machine (GBM), is able to produce competitive, highly robust, and interpretable procedures for regression and classification tasks \cite{wen2019multi}. In this paper, we use it as the representative of non-deep learning-based methods.

$\bullet$ \textbf{DNN} \cite{hinton2006fast}: We also implement a deep neural network baseline model, which has the same structure and hyper-parameters with the single branch in $ESM^2$. Different from $ESM^2$, it is trained with samples on the path ``click$\to$purchase'' or ``impression$\to$click'' to predict conversion rate $p^{cvr}$ or click-through rate $p^{ctr}$, respectively.

$\bullet$ \textbf{DNN-OS} \cite{pan2008one}: Due to the data sparsity on the paths ``impression$\to$purchase'' and ``click$\to$purchase'', it is hard to train a deep neural network with good generalization. To address this issue, we leverage the \emph{over-sampling} strategy to augment positive samples during training the deep model, named DNN-OS. It has the same structure and hyper-parameters with the above DNN model.

$\bullet$ \textbf{ESMM} \cite{ma2018entire}: For a fair comparison, we use the same backbone structure as the above deep models for ESMM. It directly models the conversion rate on the user sequential path ``impression$\to$click$\to$purchase'' without considering the purchase-related post-click behaviors.

In a nutshell, the first three methods learn to predict $p^{ctr}$ and $p^{cvr}$ using samples on the path ``impression$\to$click'' and ``click$\to$purchase'', respectively, then multiply them together to derive the click-through conversion rate $p^{ctcvr}$. As for ESMM and our $ESM^2$, they directly predict $p^{ctcvr}$ and $p^{cvr}$ by modeling them over the entire space.

\subsubsection{Hyper-parameters settings}
\label{subsubsec:implDetails}

For the GBDT model, the number of trees, depth, minimum instance numbers for splitting a node, the sampling rate of the training set for each iteration, the sampling rate of features for each iteration, and the type of loss function, are set as 150, 8, 20, 0.6, 0.6, and \emph{logistic loss}, respectively, which are chosen according to the AUC score on the validation set. For the deep neural network-based models, they are implemented in TensorFlow using the Adam optimizer. The learning rate is set to 0.0005 and the mini-batch size is set to 1000. Logistic loss is used as the loss function for each prediction task in all models. There are 5 layers in the \emph{MLP}, where the dimension of each layer is set to 512, 256, 128, 32, and 2, respectively, as summarized in Table~\ref{tab:parameter}.

\begin{table}[htbp]
  \caption{Hyper-parameters of deep neural network-based models including DNN, DNN-OS, ESMM, and $ESM^2$. }
  \label{tab:parameter}
  \begin{tabular}{ccl}
    \toprule
    Hyper-parameter & Choice\\
    \midrule
    Loss function& Logistic Loss\\
    Optimizer & Adam \\
    Number of layers in MLP & 5\\
    Dimensions of layers in MLP & [512,256,128,32,2] \\
    Batch size & 1000\\
    Learning rate& 0.0005 \\
    Dropout ratio& 0.5 \\
  \bottomrule
\end{tabular}
\label{tab:parameter}
\end{table}

\subsection{Main results}
\label{subsec:mainResults}

\subsubsection{Comparison on the offline dataset}
\label{subsubsec:offlineComparison}

In this subsection, we report the AUC, GAUC, and $F_{1}$ scores of all the competitors on the offline test set. Table~\ref{tab:auc} summarizes the results of AUC and GAUC. It can be seen that the DNN method achieves gains of 0.0242, 0.0102, 0.0117 for CVR AUC, CTCVR AUC, and CTCVR GAUC over the baseline GBDT model, respectively. It demonstrates the strong representation ability of deep neural networks. Different from the vanilla DNN, DNN-OS utilizes an over-sampling strategy to address the DS issue, achieving a better performance than DNN. As for ESMM, it is modeled on the path ``impression$\to$click$\to$purchase'', which tries to address the SSB and DS issues simultaneously. Benefiting from modeling over the entire space and the abundant training samples, it outperforms DNN-OS. Nevertheless, ESMM, neglecting the impact of post-click behaviors while being further exploited by the proposed $ESM^2$, still struggles to address the DS issue due to rare purchase training samples. After predicting some decomposed sub-targets in parallel under a multi-task learning framework, $ESM^2$ composes them sequentially to formulate the final CVR. As can be seen, it obtains the best scores among all the methods. For example, the gains over ESMM are 0.0088, 0.0101, 0.0145 for CVR AUC, CTCVR AUC, and CTCVR GAUC, respectively. It is worth mentioning that a gain of 0.01 in offline AUC always means a significant increment in revenue for online RS \cite{ma2018entire, wen2019multi}.

\begin{table}[htbp]
\label{tab:auc}
\caption{The AUC and GAUC scores of all methods.}
\begin{tabular}{c|c|c|c}
\hline
Method & CVR AUC&CTCVR AUC&CTCVR GAUC \\
\hline
GBDT & 0.7823 & 0.8059 & 0.7747\\
\cline{1-4}
DNN & 0.8065 & 0.8161 & 0.7864\\
\cline{1-4}
DNN-OS & 0.8124 & 0.8192 & 0.7893\\
\cline{1-4}
ESMM & 0.8398 & 0.8270 & 0.7906\\
\hline
$ESM^2$ & \textbf{0.8486} & \textbf{0.8371} & \textbf{0.8051} \\
\hline
\end{tabular}
\label{tab:auc}
\end{table}

\begin{table*}[htbp]
\label{tab:all_method_cvr}
\caption{The Precision, Recall and $F_{1}$ scores of all methods for CVR. }
\begin{tabular}{c|ccc|ccc|ccc}
\hline
 & \multicolumn{3}{c|}{CVR@top0.1\%} & \multicolumn{3}{c|}{CVR@top0.6\%} & \multicolumn{3}{c}{CVR@top1\%} \\
\cline{1-10}
Method & Recall & Precision & F1-Score & Recall & Precision & F1-Score & Recall & Precision & F1-Score \\
\hline
GBDT&4.382\%&14.348\%&6.714\%&16.328\%&9.894\%&12.322\%&27.384\%&7.384\%&    11.631\% \\
\cline{1-10}
DNN&4.938\%&15.117\%&7.445\%&17.150\%&10.495\%&13.021\%&28.481\%&8.196\%    &12.729\% \\
\cline{1-10}
DNN-OS&5.383\%&15.837\%&8.034\%&17.381\%&10.839\%&13.353\%&29.032\%&8.423\%    &13.058\% \\    
\cline{1-10}
ESMM&5.813\%&16.295\%&8.570\%&18.585\%&11.577\%&14.267\%&29.789\%&8.961\%    &13.777\% \\            
\hline
$ESM^2$ & \textbf{6.117\%}&\textbf{17.145\%}&\textbf{9.017\%}&\textbf{23.492\%}&\textbf{10.574\%}&\textbf{14.584\%}&\textbf{30.032\%}&\textbf{9.034\%}&\textbf{13.890\%} \\
\hline
\end{tabular}
\label{tab:all_method_cvr}
\end{table*}

\begin{table*}[htbp]
\label{tab:all_method_ctcvr}
\caption{The Precision, Recall and $F_{1}$ scores of all methods for CTCVR. }
\begin{tabular}{c|ccc|ccc|ccc}
\hline
 & \multicolumn{3}{c|}{CTCVR@top0.1\%} & \multicolumn{3}{c|}{CTCVR@top0.6\%} & \multicolumn{3}{c}{CTCVR@top1\%} \\
\cline{1-10}
Method & Recall & Precision & F1-Score & Recall & Precision & F1-Score & Recall & Precision & F1-Score \\
\hline
GBDT&2.937\%    &0.701\%    &1.132\%    &4.870\%    &0.649\%    &1.145\%    &8.894\%    &0.531\%    &1.002\% \\
\cline{1-10}
DNN&3.168\%    &0.851\%    &1.341\%    &5.269\%    &0.768\%    &1.340\%    &9.461\%    &0.643\%    &1.204\% \\
\cline{1-10}
DNN-OS&3.382\%    &0.871\%    &1.385\%    &5.369\%    &0.801\%    &1.395\%    &9.863\%    &0.673\%    &1.260\% \\    
\cline{1-10}
ESMM&3.858\%&0.915\%    &1.479\%    &5.504\%    &0.828\%    &1.439\%    &10.088\%    &0.691\%    &1.294\%     \\
\hline
$ESM^2$ & \textbf{4.219\%}&\textbf{1.001\%}&\textbf{1.618\%}&\textbf{5.987\%}&\textbf{0.900\%}&\textbf{1.566\%}&\textbf{10.991\%}&\textbf{0.753\%}&\textbf{1.410\%}\\
\hline
\end{tabular}
\label{tab:all_method_ctcvr}
\end{table*}

As for the $F_{1}$ score, we report several values by setting different thresholds for CVR and CTCVR, respectively. First, we sort all the instances in descending order according to the predicted CVR or CTCVR score. Then, due to the sparsity of CVR task (about 1\% of the predicted samples are positive), we choose three thresholds namely top@0.1\%, top@0.6\%, and top@1\% to split the predictions into positive and negative groups accordingly. Finally, we calculate the precision, recall, and $F_{1}$ scores of these predictions at these different thresholds. Results are summarized in Table~\ref{tab:all_method_cvr} and Table~\ref{tab:all_method_ctcvr}. A similar trend to Table~\ref{tab:auc} can be observed. Again, the proposed method $ESM^2$ achieves the best performance in different settings.

\subsubsection{Comparison on online deployment}
\label{subsubsec:onlineComparison}
\begin{figure}[htbp]
\includegraphics[width=1\linewidth]{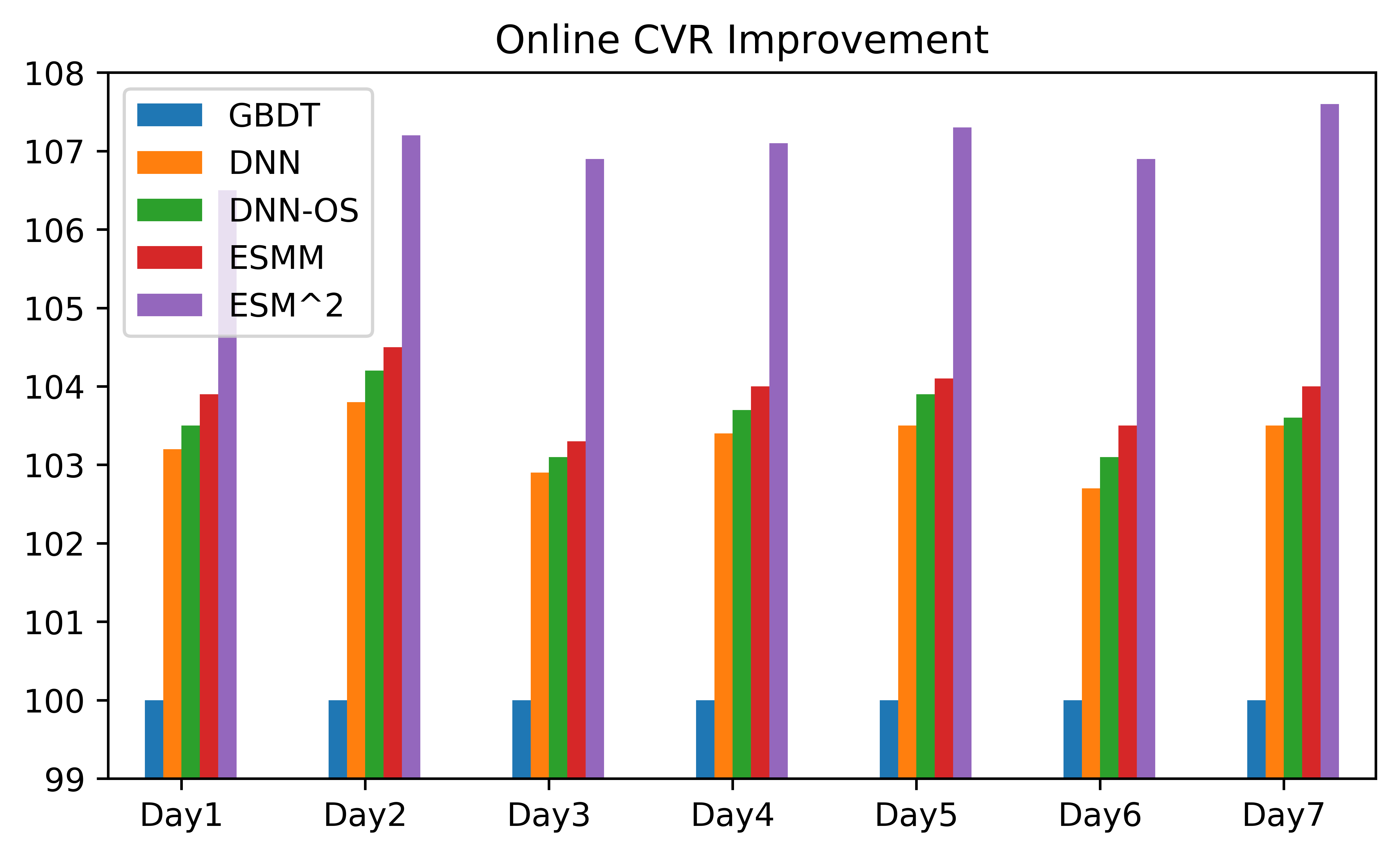}
\caption{The results of A/B test for CVR by deploying different models in our recommender system. }
\label{fig:ab_test}
\end{figure}

It is not an easy job to deploy deep network models in our recommender system since it servers hundreds of millions of users every day, $e.g.$, more than 100 million users per second at a traffic peak. Therefore, a practical model is required to make real-time CVR prediction with high throughput and low latency. For example, hundreds of recommendation items for each visitor should be predicted in less than 100 milliseconds in our system. Thanks to the parallel network structure, our model is computationally efficient and can respond to each online request within 20 milliseconds. To make the online evaluation fair, confident, and comparable, each deployed method for an A/B test has involved the same number of users, $i.e.$, millions of users. The results are listed in Figure~\ref{fig:ab_test}, where we use the GBDT model as the baseline. As can be seen, DNN, DNN-OS, and ESMM achieve comparable performance and outperform the baseline model significantly, while ESMM performs slightly better. As for the proposed $ESM^2$, the significant margins between it and the above methods demonstrate its superiority. Besides, it contributes to a 3\% CVR promotion compared with ESMM, indicating a significant business value for the e-commercial platform.

\textbf{Remarks:} 1) The deep neural network has stronger representation ability than the decision tree-based GBDT; 2) the multi-task learning framework over the entire sample space serves as an efficient tool to address the SSB and DS issues; and 3) based on the idea of post-click behaviors decomposition, $ESM^2$ efficiently addresses the SSB and DS issues by modeling CVR over the entire space and leveraging abundant supervisory signals from deterministic actions and achieves the best performance.

\subsection{Ablation studies}
\label{subsec:ablation}
In this part, we present the detailed ablation studies including hyper-parameter settings of the deep neural network, effectiveness of embedding dense numerical features, and the choice of decomposing post-click behaviors, respectively.

\begin{figure*}
\includegraphics[width=1\linewidth]{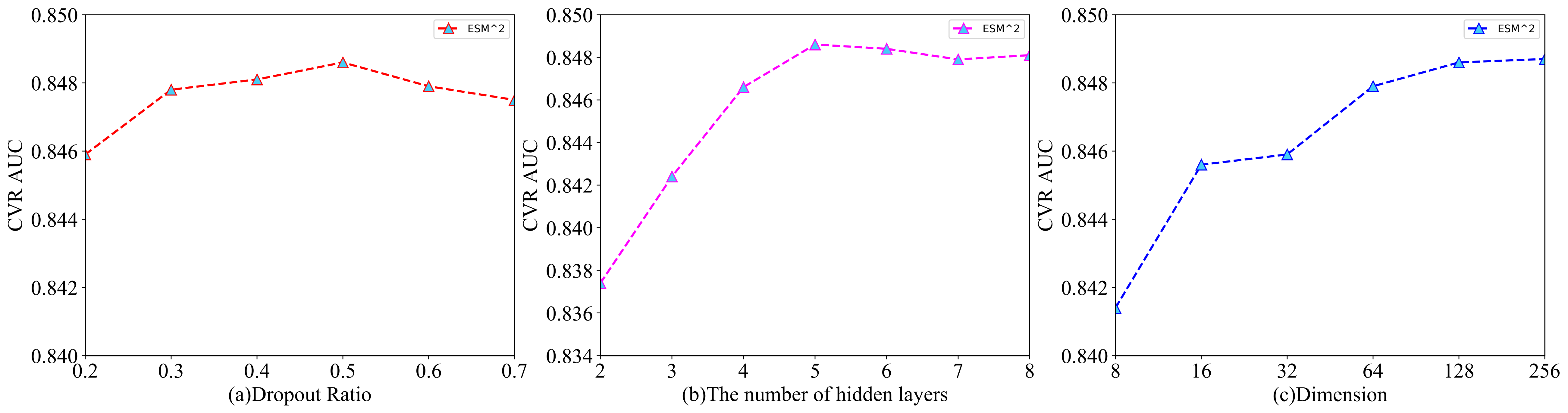}
\caption{The results of different hyper-parameter settings in $ESM^2$. }
\label{fig:dimension}
\end{figure*}

\subsubsection{Hyper-parameters of deep neural network}
\label{subsubsec:hyperParamsDNN}
Here, we take three critical parameters, namely \emph{dropout ratio}, \emph{the number of hidden layers}, and \emph{the dimension of item feature embeddings} as example to illustrate the process of parameter selection in our $ESM^2$ model. 

Dropout \cite{srivastava2014dropout} refers to the regularization technique by randomly deactivating some neural nodes during training. It can increase the generalization of the deep neural network by introducing randomness. We try different choices of the dropout ratio from 0.2 to 0.7 in our model. As shown in Figure~\ref{fig:dimension}(a), a dropout ratio of 0.5 leads to the best performance. Therefore, we set it to 0.5 in all the experiments if not specified.

Increasing the depth of network layers can enhance the model capacity but also potentially leads to over-fitting. Therefore, we carefully set this hyper-parameter according to the AUC scores on the validation set. As can be seen from Figure~\ref{fig:dimension}(b), at the beginning stage, $i.e.$, from two layers to five layers, increasing the number of hidden layers consistently improves the model's performance. However, it saturates at five layers that increasing more layers even marginally decreases the AUC scores, where the model may overfit the training set. Therefore, we use five hidden layers in all experiments if not specified.

The dimension of item feature embeddings is a critical parameter that high-dimension features reserve more information but also contains noise and leads to higher model complexity. We try different settings of the parameter and plot the results in Figure~\ref{fig:dimension}(c). As can be seen, increasing the dimension generally improves performance. It finally saturates at 128 while doubling it leads no more gains. Therefore, to make a trade-off between model capacity and complexity, we set the dimension of item feature embeddings to 128 in all the experiments if not specified.

\subsubsection{Effectiveness of embedding dense numerical features} 
In our task, there are several numerical features. A common practice is to discretize them into one-hot vectors first and then concatenate them with ID features together, which are then embedded into dense features through a linear projection layer as described in Section~\ref{subsec:ESM}. However, we hypothesize that the one-hot vector representation of numerical features may degrade the precision during discretization. In contrast, we try another solution by normalizing the numerical features and embedding them using the Tanh activation function, $i.e.$,
\begin{equation}
   {g_{ij}} = \tanh \left( {\frac{{{f_{ij}} - {\mu _{{f_j}}}}}{{{\sigma _{{f_j}}}}}} \right),
   \label{eq:normalization}
\end{equation}
where ${{\mu _{{f_j}}}}$ and ${{\sigma _{{f_j}}}}$ denotes the mean and standard deviation of the $j^{th}$ kind of features. Then, we concatenate the embedded features with the ID features together as the input of our $ESM^2$ model. It achieves a gain of 0.004 AUC over the discretization-based method. Therefore, we use the normalization-based embedding method for dense numerical features in all the experiments if not specified.

\subsubsection{Effectiveness of decomposing post-click behaviors} 
When decomposing the post-click behaviors, we can integrate different behaviors into the DAction node, $e.g.$, only SCart, only Wish, or both SCart and Wish (SCart and Wish). Here, we evaluate the effectiveness of different choices. The results are summarized in Table~\ref{tab:singal_auc}. As can be seen, the combination of both SCart and Wish achieves the best AUC scores. It is reasonable since there are more purchase-related labeled data than the other two cases to address the DS issue. 

\begin{table}[h]
\label{tab:singal_auc}
\caption{The results of choices on post-click behaviors.}
\begin{tabular}{c|c|c|c}
\hline
 & CVR AUC & CTCVR AUC & CTCVR GAUC \\
\hline
SCart & 0.8457 & 0.8359 & 0.7996\\
\cline{1-4}
Wish & 0.8403 & 0.8319 & 0.7962 \\
\cline{1-4}
SCart and Wish & \textbf{0.8486} & \textbf{0.8371} & \textbf{0.8051} \\
\hline
\end{tabular}
\label{tab:singal_auc}
\end{table}

\subsection{Performance analysis of user behaviors}
\label{subsec:performanceAnalysis}
\begin{figure}
\includegraphics[width=1\linewidth]{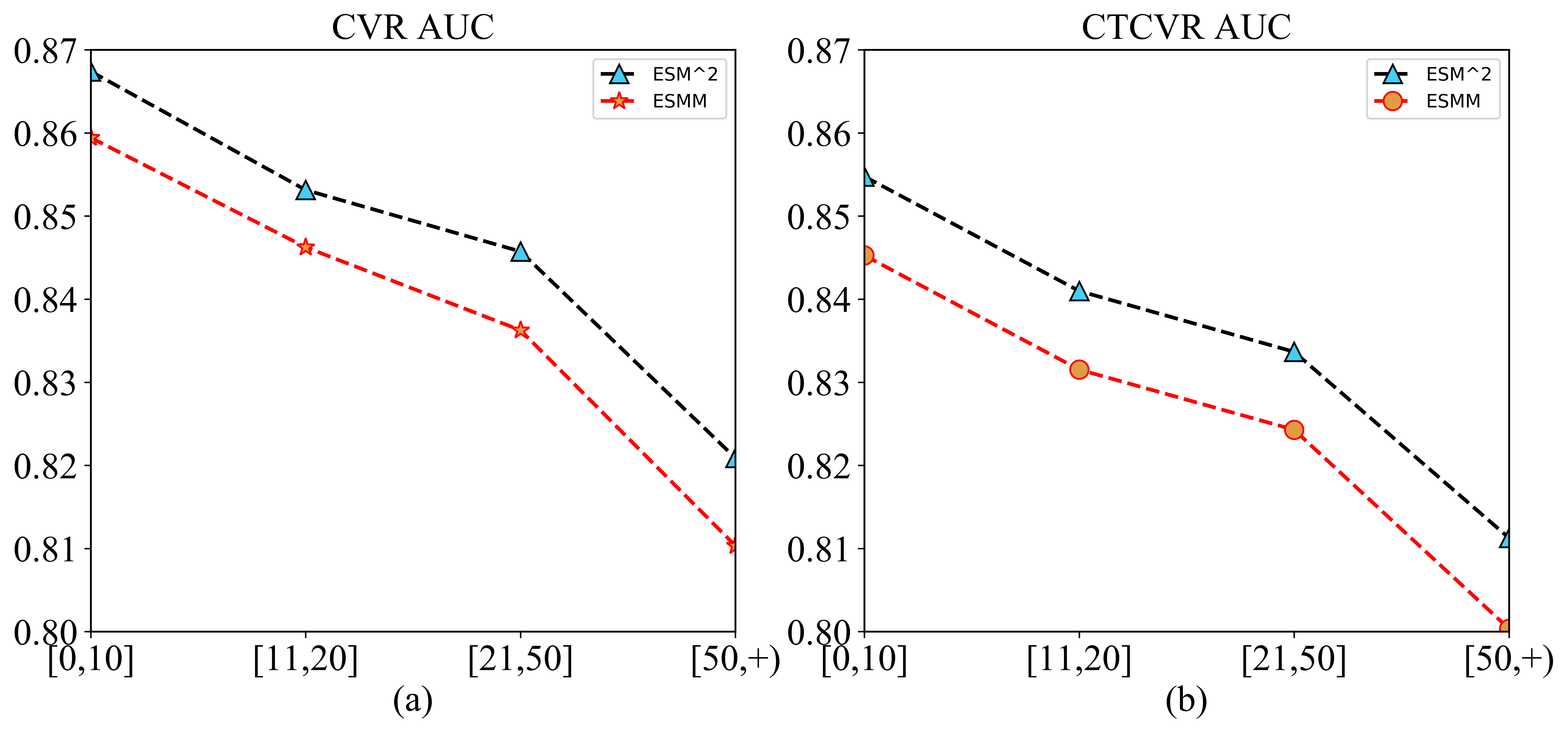}
\caption{The AUC scores of CVR and CTCVR for ESMM and $ESM^2$ at different groups regarding the number of purchasing behaviors. Please refer to Section~\ref{subsec:performanceAnalysis}. }
\label{fig:esmm_esm_user}
\end{figure}

To understand the performance of $ESM^2$ and its difference with ESMM, we further partition the test set into four groups according to the number of users' purchasing behaviors, $i.e.$, [0,10], [11,20], [21,50], [50,+). We report AUC scores of CVR and CTCVR for both methods at each group, and the results are plotted in Figure~\ref{fig:esmm_esm_user}. As can be seen, the CVR AUC(CTCVR AUC) of both methods decreases with the increase of the number of purchasing behaviors. However, we observe that the relative gain of $ESM^2$ over ESMM in each group increases, $i.e.$, 0.72\%, 0.81\%, 1.13\%, 1.30\%. Generally, users having more purchasing behaviors always have more active post-click behaviors such as SCart and Wish. Our $ESM^2$ model deals with such post-click behaviors by adding a DAction node supervised by deterministic signals from users' feedback. Therefore, it has better representation ability on those samples than ESMM and achieves better performances on the users with high-frequency purchasing behaviors.

\section{Conclusion}
\label{sec:Conclusion}
In this paper, we introduce a novel idea of post-click behavior decomposition for modeling CVR task in the context of e-commerce recommender system. A novel user sequential behavior graph ``impression$\to$click$\to$D(O)Action$\to$purchase'' is constructed, which is used to model CVR over the entire space. Based on the conditional probability rule, we disentangle CVR and some related auxiliary tasks including the post-view click-through rate, click-through DAction conversation rate, and click-through conversion rate into four hidden probability variables, which are defined on explicit sub-paths of the graph. Consequently, we propose a novel deep neural recommendation model named $ESM^2$ by employing a multi-task learning framework to predict CVR as well as related auxiliary tasks simultaneously. By training with all impression samples and leveraging the abundant labels of deterministic post-click behaviors, our $ESM^2$ model efficiently addresses the SSB and DS issues. Extensive experiments on both offline and online environments demonstrate the superiority of $ESM^{2}$ over state-of-the-art models.

\section*{Acknowledgment}
This work was partly supported by the National Natural Science Foundation of China (NSFC) under Grant 61806062.

\bibliographystyle{ACM-Reference-Format}
\bibliography{2020SIGIR_RS}

\end{document}